\documentclass[letterpaper, 10 pt, conference]{ieeeconf}

\IEEEoverridecommandlockouts                              

\usepackage[space, compress, sort]{cite}

\usepackage{amsmath, amssymb, amsfonts, mathtools, amsthm}

\usepackage{graphicx}
\usepackage{textcomp}
\usepackage{xcolor}
\def\BibTeX{{\rm B\kern-.05em{\sc i\kern-.025em b}\kern-.08em
    T\kern-.1667em\lower.7ex\hbox{E}\kern-.125emX}}
    
\usepackage[linesnumbered,ruled]{algorithm2e}
\usepackage{xcolor}
\usepackage{paralist}
\usepackage[caption=false]{subfig}

\usepackage{bm}
\usepackage{scalerel}
\usepackage{multirow}
\usepackage{array, makecell} %
\usepackage{adjustbox}

\theoremstyle{plain}
   
\theoremstyle{remark}

\theoremstyle{definition}

\theoremstyle{plain}

\newcommand{\todo}[1]{{\color{red}{#1}}}

\DeclarePairedDelimiterX{\norm}[1]{\lVert}{\rVert}{#1}
    
\usepackage[para,online,flushleft]{threeparttable}
\usepackage{array,booktabs,makecell}
\usepackage{bm}
\let\labelindent\relax
\usepackage[inline]{enumitem}
\usepackage{booktabs}
\usepackage{relsize}
\usepackage{accents}
\usepackage{array}
\usepackage{tabularx}
\usepackage{multirow}

\usepackage{tcolorbox}
\usepackage{listings}
\usepackage{makecell}
\newcommand{\NA}{---} 

\let\oldnl\nl
\newcommand{\nonl}{\renewcommand{\nl}{\let\nl\oldnl}}

\definecolor{lightgray}{rgb}{0.96, 0.96, 0.98}
\definecolor{lightyellow}{rgb}{0.98, 0.97, 0.97}

\usepackage{algorithmic}
\usepackage[linesnumbered,ruled]{algorithm2e}

\title{\LARGE \bf PIP-LLM: Integrating PDDL-Integer Programming with LLMs for Coordinating  Multi-Robot Teams Using Natural Language
}

\author{ Guangyao Shi$^{1}$\textsuperscript{\textdagger}, Yuwei Wu$^{2}$, Vijay Kumar$^{2}$, Gaurav S. Sukhatme$^{1}$ 
\thanks{$^{1}$Guangyao Shi and Gaurav S. Sukhatme are with with the Department of Computer Science, University of Southern California, Los Angeles, CA 90089, USA. Email:
\texttt{\small \{shig, gaurav\}@usc.edu}.}
\thanks{$^{2}$Yuwei Wu and Vijay Kumar are with the GRASP Lab, University of Pennsylvania, Philadelphia, PA 19104, USA. Email: \texttt{\small \{yuweiwu, kumar\}@seas.upenn.edu}.}
\thanks{\textsuperscript{\textdagger} Corresponding author.}
}
\begin{document}
\maketitle
\thispagestyle{empty}
\pagestyle{empty}

\begin{abstract}
  Enabling robot teams to execute natural language commands requires translating high-level instructions into feasible, efficient multi-robot plans. While Large Language Models (LLMs) combined with Planning Domain Description Language (PDDL) offer promise for single-robot scenarios, existing approaches struggle with multi-robot coordination due to brittle task decomposition, poor scalability, and low coordination efficiency. 
  We introduce PIP-LLM, a language-based coordination framework that consists of PDDL-based team-level planning and Integer Programming (IP) based robot-level planning. PIP-LLMs first decomposes the command by translating the command into a team-level PDDL problem and solves it to obtain a team-level plan, abstracting away robot assignment. Each team-level action represents a subtask to be finished by the team. Next, this plan is translated into a dependency graph representing the subtasks' dependency structure. Such a dependency graph is then used to guide the robot-level planning, in which each subtask node will be formulated as an IP-based task allocation problem, explicitly optimizing travel costs and workload while respecting robot capabilities and user-defined constraints. This separation of planning from assignment allows PIP-LLM to avoid the pitfalls of syntax-based decomposition and scale to larger teams. Experiments across diverse tasks show that PIP-LLM improves plan success rate, reduces maximum and average travel costs, and achieves better load balancing compared to state-of-the-art baselines.
\end{abstract}

\section{Introduction}
A long-standing goal in multi-robot research is to create systems that take concise human language commands and automatically generate executable coordination plans for robot teams. Achieving this level of autonomy requires accurate interpretation of human commands, decomposition of these commands into executable subtasks, and effective allocation of tasks across robots~\cite{rizk2019cooperative}. Recent advances in large language models (LLMs) offer a promising path toward this objective.

LLMs, trained on large-scale text corpora, exhibit common-sense reasoning abilities and can process a wide range of tasks expressed in natural language. They have achieved notable success in domains such as robotic manipulation~\cite{liu2024enhancing, maeureka}, navigation~\cite{yu2023l3mvn}, and locomotion~\cite{tang2023saytap}. Some studies have directly employed LLMs as task planners, showing moderate success in single-robot sequencing problems. However, in more complex scenarios, LLMs alone often produce suboptimal or infeasible solutions~\cite{kannan2024smart, mandi2024roco}, which is especially problematic in multi-robot coordination where feasibility and optimality are equally critical. Consequently, many recent works integrate LLMs with classical planning~\cite{ghallab2004automated}, for example, by translating natural language (NL) commands into Planning Domain Description Language (PDDL) and leveraging existing solvers~\cite{liu2025delta, liu2023llm+}. Yet, most of these approaches remain limited to single-robot domains, and extending them to multi-robot contexts introduces additional challenges. First, formulating multi-robot problems into a single structured representation is significantly more complex, and LLMs frequently produce erroneous or incomplete encodings (“hallucinations”~\cite{Li_2025_CVPR, ji2023towards, li2025videohallu, li2025self, guan2024hallusionbench}) when handling large teams and complex tasks. Second, even with correct formulations, standard solvers for PDDL often fail to scale to the large state and action spaces of multi-robot planning. Third, numeric objectives and constraints (e.g., energy minimization or resource-constrained task allocation) are difficult to specify in PDDL, and the corresponding solvers remain underdeveloped.

\begin{figure*}
\centerline{\includegraphics[scale=0.95]{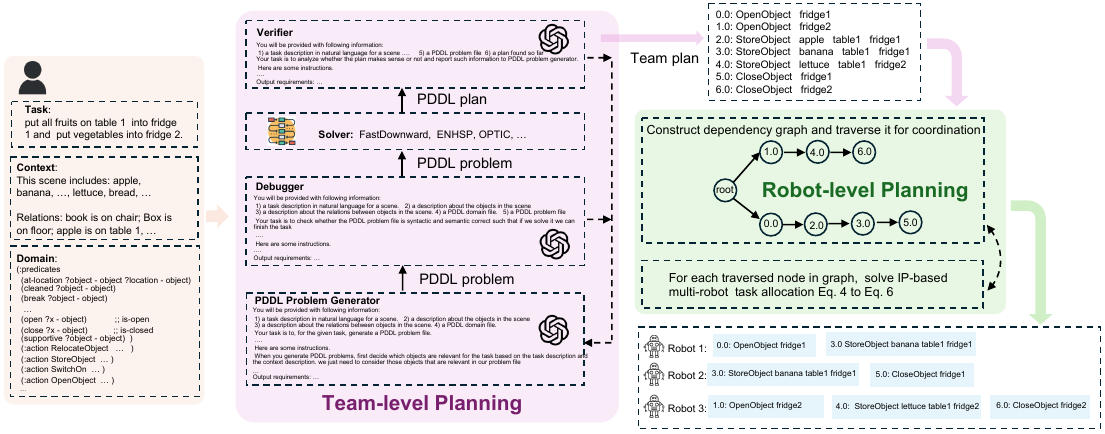}}
\caption{Overview of PIP-LLM. Users provide the task, context, and PDDL domain description in natural language (orange box). Then, LLMs will analyze the relevant information to construct a PDDL problem instance (\textit{PDDL Problem Generator}). Next, the result is fed to a \textit{Debugger} and a \textit{Verifier}, which will check the syntax of PDDL problems and the correctness of the solution and feed the information back to the problem generation LLM. The generation process may be repeated several rounds. Each action in the output plan corresponds to a subtask that the robots as a team can do. Then, PIP-LLM uses this plan to construct a dependency graph to identify the dependencies between subtasks and facilitate parallel execution, which will then be used to allocate and schedule robots to finish all sub-tasks (green box, robot-level planning).}
\vspace{-0.3cm}
\label{fig:framework}
\end{figure*}

Several efforts have attempted to extend the LLM+classic planning paradigm to multi-robot systems. Bai et al.~\cite{bai2024twostep} propose decomposing PDDL goals into robot-specific subgoals to decouple planning across robots. However, their method does not generalize well to teams with more than two robots or scenarios where decomposition is non-trivial. Zhang et al.~\cite{zhang2024lamma} introduce {LaMMA-P}, which extends the architecture proposed by Kannan et al.~\cite{kannan2024smart} by incorporating PDDL planning for each robot. Both frameworks rely heavily on pure syntactic segmentation and analysis of natural language commands for task decomposition, which is inherently unreliable. For instance, suppose the command is \texttt{"put five apples in the fridge"},  the naive decomposition based on syntactic segmentation and analysis is to decompose the task into five subtasks, each of which is to put an apple in the fridge. However, some robots may be capable of moving two apples at a time, and some robots move one. 
How to decompose the task depends not only on textual information but also on the robots’ capabilities and their availability.  
Similarly, a command such as \texttt{"put all fruits and vegetables on the table into fridges 1 and 2"} requires contextual reasoning about object quantities and container capacities, which cannot be captured by naive language segmentation. These limitations hinder their applicability to large-scale, heterogeneous, or long-horizon tasks. Moreover, their frameworks implicitly assume a special case of multi-robot task allocation (MRTA) \cite{gerkey2004formal} where the number of subtasks is no greater than the number of robots and allocation is determined solely by skill requirements without explicitly considering task costs (e.g., travel distance) and the transition costs between tasks.  Generally,  however, the number of subtasks may exceed the number of robots in MRTA; tasks may vary in duration; and robots may need to transit between multiple tasks. Without explicitly considering task costs, solutions can be highly suboptimal; without proper coordination, workload imbalances lead to large travel costs of a particular robot, i.e., some robots are busy all the time while others keep idling.   

To address these challenges, we propose a novel framework, \textbf{PIP-LLM}, for language-driven multi-robot coordination (Fig.~\ref{fig:framework}). The key idea is to form a hierarchical planning structure: using LLM+PDDL for team-level task understanding and then using Integer Programming (IP) for robot assignments. Specifically, given a natural language command, our framework first formulates a team-level PDDL problem that captures what the robot team as a whole must accomplish, independent of specific robot assignments. The resulting plan is then represented as a dependency graph by analyzing action dependencies, which will then be used to schedule the robotic team. Unlike prior work that relies on syntactic segmentation, this approach provides semantically grounded and formal task decomposition by formulating and solving PDDL problems. In the second stage, a skill-based allocation framework using IP assigns tasks to robots based on their capabilities and task costs, ensuring both feasibility and efficiency. 
This two-level design enables the framework to robustly handle complex commands, scale efficiently to large numbers of robots and objects, and achieve more balanced and near-optimal execution.
Furthermore, user preferences can be seamlessly incorporated into task allocation; for example, constraints expressed in natural language, such as excluding a particular robot from certain tasks.

The main contributions of this paper are as follows:
\begin{itemize}
    \item We propose a novel multi-robot task planning framework that integrates PDDL-based team-level planning with IP-based robot-level allocation, enabling flexible and efficient coordination of robot teams using natural language commands. 
    \item We introduce a benchmark dataset for evaluating multi-robot task planning in large-scale, long-horizon scenarios involving many robots and objects. 
    \item We provide extensive simulation results and demonstrate real-world applications to validate the effectiveness and scalability of the proposed framework.
\end{itemize}

\section{Related Work}\label{sec:rel_wrk}

Recent advances have demonstrated the potential of integrating LLMs with individual robotic platforms to enable natural human–robot interaction and high-level task planning. 
By leveraging the strengths of LLMs in language understanding and contextual reasoning, these systems translate free-form instructions into executable action sequences. 
Because natural language is inherently contextual and semantic, LLMs are well-suited for navigation, exploration, and instruction following; in such settings, they help ground linguistic inputs into spatial and semantic representations that guide behavior in complex, partially observed environments~\cite{10.1609/aaai.v38i17.29858, ravichandran_spine, shah2022lmnav}. In addition, modules for real-time residency~\cite{10885890} and anomaly detection~\cite{sinha2024real} have been integrated into LLM-based robotic pipelines to improve robustness in dynamic and unforeseen conditions.

Built on this, researchers are increasingly exploring how LLMs can facilitate coordination for multi-robot teams~\cite{xu2024nl2hltl2planscalingnaturallanguage, mandi2024roco, chen2024scalable, wang2024probabilistically, yu2023co, kannan2024smart}. Some pioneering works seek to design prompt frameworks to solve multi-robot coordination problems using mainly LLMs~\cite{mandi2024roco, kannan2024smart}. Zhao et al. \cite{mandi2024roco} proposed a dialogue-based framework to coordinate multiple robots.  Such approaches are not suitable to solve long-horizon
and large-scale multi-robot problems because such problems will require robots to communicate many rounds with redundant (possibly misleading as the number of rounds increases) information. 
LLMs tend to get lost and hallucinate. Chen et al. \cite{chen2024scalable} did an extensive evaluation of communication frameworks (centralized, decentralized, or hybrid) to find a scalable solution for multi-robot coordination. Another representative work along this direction is \cite{kannan2024smart}, in which the authors propose a framework for multi-robot task planning based on the canonical decomposition of multi-robot task planning, i.e., task decomposition, coalition formation, and
task allocation 
\cite{rizk2019cooperative}. 
Such a framework does not use any external tools and purely relies on LLMs' reasoning capability to solve problems. 
However, it cannot be used to coordinate large-scale robot teams, and it does not consider the optimality aspects of the coordination.

Another line of work uses LLMs to translate natural language into structured intermediate representations \cite{xu2024nl2hltl2planscalingnaturallanguage,obata2024lip,lykov2023llm,zhang2024lamma}. Xu et al. \cite{xu2024nl2hltl2planscalingnaturallanguage} use hierarchical temporal logic as an intermediate representation for temporally dependent multi-robot task planning. Zhang et al. \cite{zhang2024lamma} extend \cite{kannan2024smart} by adding a PDDL representation for each robot's individual planning. 
Our work aligns with this direction.
The proposed framework adopts a hierarchical structure that employs two distinct representations at different levels: PDDL for team-level planning and IP for robot-level planning.
Our framework can cover both the feasibility and optimality aspects of multi-robot coordination and is very suitable for long-horizon tasks that involve a large number of robots.

\section{Problem Formulation}
\subsection{Preliminaries}
\noindent \textbf{PDDL}: As a standard, declarative language for specifying automated planning tasks~\cite{haslum2019introduction, fox2003pddl2}, it separates the domain, which specifies types, predicates, and parameterized actions with preconditions and effects, from the problem, which enumerates concrete objects, an initial state, and goals.
Beyond the basic STRIPS subset, PDDL supports richer constructs such as conditional and quantified effects, numeric fluents, and durative actions.

\noindent \textbf{Robot Model}: 
We consider a set of $N$ (possibly) heterogeneous robots, $\mathcal{R}=\{R^{1}, R^{2},\ldots, R^{N}\}$. Let $\Delta$ denote the set of all skills or actions that a robot may be capable of performing. We assume that skills in $\Delta$ are either pre-implemented in the system or accessible through available API calls. Each agent $R^{n}$ possesses a subset of skills, $S^{n}\subseteq \Delta$, subject to specific constraints. For example, the robot skill \texttt{PickUpObject} may be limited by the maximum mass that the robot can lift.

\noindent \textbf{Environment Model}: We assume a static and deterministic environment. The state of the environment can be explicitly represented. The number of objects and robots is fixed, and the robot's action will have a deterministic effect on the state of the environment.

\subsection{Problem Statement}
Given a high-level natural-language instruction \(I\), our aim is to parse the instruction, decompose it into the requisite subtasks, and synthesize an executable plan. Subtasks are scheduled to maximize robot utilization, enabling parallel execution when precedence and resource constraints allow. Environment \(E\) contains multiple robots and objects, and \(I\) is assumed to be feasible within \(E\).

\section{Proposed Framework}
\subsection{Overview}
The proposed framework is shown in Fig. \ref{fig:framework}, which hierarchically conducts task planning for multi-robot systems.

\noindent \textbf{Team-Level Planning}. Given a task description in natural language, LLMs will first analyze the context information, relevant predicates, and goals that need to be achieved. Based on this analysis, it will construct a PDDL problem instance. Then, the problem instance will be fed to a \text{Debugger} and a \textit{Verifier}, which will check the syntax of PDDL problems and the correctness of the solution and feed the information back to the problem generation LLM. The generation process will be repeated until the set loop limit is reached.   Each action in the output plan corresponds to a subtask that the robots as a team can do. Then, we will use this plan to construct a dependency graph, which helps us to identify the dependencies between subtasks and facilitate parallel execution. Details on how to construct such 
task trees will be discussed in Sec. \ref{sec:construct_task_tree}. The team-level planning part of PIP-LLM is, in essence, doing task decomposition.  But it is fundamentally different from those in~\cite{zhang2024lamma, kannan2024smart} which rely on syntactic language segmentation to decompose tasks. Instead, our framework uses LLMs to construct a formal high-level plan and represent tasks as a structured format, a dependency tree.  It should be noted that our team-level PDDL planning problem is different from the robot-specific PDDL planning problem ~\cite{zhang2024lamma, bai2024twostep} in the sense that the action is about the team as a whole can without referring to any specific robots. Such an abstract reduces the information fed to LLMs and allows them to do high-level reasoning more accurately.

\begin{algorithm}[!ht]
    \caption{Dependency Graph Scheduling}
    \label{algorithm:graph_scheduling}
    \SetKwInOut{Input}{Input}
    \SetKwInOut{Output}{Output}
    \SetKwProg{Fn}{Function}{:}{}
    \SetKwFunction{DFS}{DFS\_visit}
    \Input{
         A subtask dependency graph $\mathcal{G}=(\mathcal{V}, \mathcal{E})$
    }
    \Output{
    Ordered robot actions
    }
     create empty stack $H$   \quad \textit{\# stack for BFS frontier} \\
     $H$.insert({$root$}) \\
     
    \While{schedule is triggered and $H$ is not empty}{
        current\_subtasks $\gets$ $H$.next\_task() \\
        \texttt{sol} $\gets$ \texttt{skill\_MRTA} (current\_subtasks)  \\
        allocate subtasks to robots using \texttt{sol} \\
        add child nodes of current\_subtasks in $\mathcal{G}$ to $H$
    }
\end{algorithm}

\noindent \textbf{Robot-Level Planning}. The constructed dependency graph will then be used to allocate and schedule robots to finish all sub-tasks. Specifically, we will traverse the task tree in a Breadth-First-Search (BFS) fashion. We maintain a BFS frontier in a stack. The task nodes in this frontier can be executed in parallel. When we need to allocate robots for the task, we pop one node from the frontier and update the stack. More details are given in Algo. \ref{algorithm:graph_scheduling}. Whenever we traverse a node in the dependency graph, we will solve a skill-based MRTA problem to allocate the subtask to specific robots to finish. Details will be discussed in Sec. \ref{sec:MRTA}. Moreover, we may incorporate human preference constraints into the MRTA formulation to accommodate humans' extra requirements of robots.

Combining both team-level planning and robot-level planning, as we will show in the experiments, such a hierarchy planning structure in our framework can not only achieve a high success rate in fulfilling diverse tasks, but also improve the overall coordination efficiency of the multi-robot systems compared to baselines.

\subsection{From Team-Level Plan to Subtask Dependency Graph}\label{sec:construct_task_tree}

The output from the PDDL planner is a sequence of team actions. If we execute the actions in order, we can finish the tasks. However, it is not necessary to strictly follow the order because some actions in the plan may be parallelizable. An example is shown in Fig. \ref{fig:framework}.  Given the domain, we can get a team plan as shown at the upper right of Fig. \ref{fig:framework}. It should be noted that for the first action (\texttt{OpenObject fridge1}) in the plan, its effect will not affect any preconditions in the next action (\texttt{OpenObject fridge2}). Similarly, there is no conflict in preconditions and effects between \texttt{StoreObject apple table1 fridge1} and \texttt{StoreObject lettuce table1 fridge2}. In such a case, we can actually execute these two actions in parallel. Inspired by this, we propose a simple action parallelization algorithm as shown in Algo.~\ref{algorithm:graph_generation} to parallelize the team actions obtained from our team-level planning. The resulting dependency graph of the plan is also in the robot-level planning block of Fig. \ref{fig:framework}.
\begin{algorithm}[ht]
    \caption{Dependency Graph Generation}
     \label{algorithm:graph_generation}
    \SetKwInOut{Input}{Input}
    \SetKwInOut{Output}{Output}
    \Input{
    \begin{itemize}
        \item A PDDL domain file, a problem file
        \item A plan file $\mathcal{P}$ as an ordered list 
    \end{itemize}
    }
    \Output{
    A subtask dependency graph $\mathcal{G}$
    }
    $\mathcal{G} \gets$ root node\\
    \For{action\_index, action in $\mathcal{P}$}{
        \nonl \# trace back to find a proper parent node \\
        \For{$i$ in {range(action\_index-$1$, $1$, $-1$)}}{ 
            flag $\gets$ Check\_dependency(\textit{action}, $\mathcal{P}[i]$) \\
        \If{flag and no parent node added}
            {add \textit{action} as child node of $\mathcal{P}[i]$ in $\mathcal{G}$ \\break } 
        }
        \If{no parent node added}
        {add \textit{action} as child node of root of $\mathcal{G}$}
    }
    return $\mathcal{G}$
\end{algorithm}

\subsection{Skill-based Multi-Robot Task Allocation}\label{sec:MRTA}
We consider a team of \(N\) (potentially heterogeneous) robots, where each robot is characterized by a vector of continuous capability \emph{skills}~\cite{ravichandar2020strata}. The skills of robot \(i\) are
\begin{equation}
    q^{(i)} = \big[q^{(i)}_1,\, q^{(i)}_2,\, \ldots,\, q^{(i)}_U\big] \in \mathbb{R}_{\ge 0}^{\,U},
\end{equation}
where \(q^{(i)}_u \in \mathbb{R}_{\ge 0}\) denotes the level of the \(u\)-th skill. If robot \(i\) lacks skill \(u\) (e.g., a car-like robot cannot open a refrigerator door), then \(q^{(i)}_u = 0\); otherwise \(q^{(i)}_u > 0\) quantifies its capacity. For example, if a robot can pick up 2 units of objects, its pickup skill is 2.0. 
Stacking the robot skill vectors gives the team skill matrix
$ \mathbf{Q} = [q^{(1)^{T}}, \ldots, q^{(N)^{T}}] \in \mathbb{R}_{+}^{U \times N} $
whose \(i\)-th column and \(u\)-th row corresponds to robot \(i\) and  skill \(u\), respectively.

Each task is specified by the skill requirements needed to complete it. For example, the task \texttt{put a lunch box in the fridge} may require the skills \texttt{open\_door}: \(1.0\), \texttt{close\_door}: \(1.0\), \texttt{pick\_up}: \(1.2\), and \texttt{drop}: \(1.2\), where \(1.2\) encodes a weight affordance. Formally, the desired task skills are $\mathbf{Y} = \big[y_1,\, y_2,\, \ldots,\, y_U\big]^T \in \mathbb{R}_{\ge 0}^{\,U},$
with \(y_u = 0\) if skill \(u\) is unnecessary and \(y_u > 0\) otherwise.

Assigning robot \(i\) to a task incurs a cost \(c_i \in \mathbb{R}_{\ge 0}\); in this work, we take \(c_i\) to be the total  travel distance of robot \(i\).
Given the task skill requirement, robot skill matrix, and the cost, we will solve the following lexicographic multi-objective IP problem to decide robot task allocation,
\begin{subequations}
\begin{align}
    \min &\quad \{ \max_{i} c_{i}x_{i}, \quad \sum_i {c_{i}x_{i}}\} \\
    \text{s.t.}~&\mathbf{Q} \mathbf{x} \geq \mathbf{Y}, \quad  x_{i} \in \{0, 1\}, \label{eq:skill_constraints}\\
    & \texttt{customized constraints}, \label{eq:other_IP_constraint}
\end{align}
\end{subequations}
where $x_{i}$ is a binary variable: $x_{i}=1$ if robot $i$ is assigned to the task, otherwise 0; $\mathbf{x} =[x_{1}, x_{2}, \ldots, x_{N}]^T$. The first objective $\max_{i} c_{i}x_{i}$ denotes the maximum cost of the team, and the second objective $\sum_i {c_{i}x_{i}}$ denotes the sum of the team cost. Eq. \eqref{eq:other_IP_constraint} denotes other domain-specific constraints or preference constraints specified by users (an example is given in Tab. \ref{tab:warehouse}, ID=5). 
Such problems can be solved using existing solvers like Gurobi. 

\section{Experiments}
We conducted two types of simulation experiments to evaluate the performance of the proposed PIP-LLM framework. In the first type of experiment, we use the AI2THOR simulator~\cite{ai2thor} as \cite{kannan2024smart, zhang2024lamma}, which is designed for domestic tasks. The goal of this type of experiment is to evaluate whether the proposed framework can improve the team coordination efficiency in terms of travel costs while not compromising the success rate. In the second type of experiment, we used a customized warehouse environment in Gazebo, which involved coordinating many robots to rearrange many objects. The goal of the second type of experiment is to show the scalability and flexibility of the proposed framework in coordinating a large number of robots with concise language commands. Moreover, we validate our results with a hardware demonstration. 

\begin{table*}[t]
\centering
\caption{Evaluation of PIP-LLM and baselines in the AI2-THOR simulator.}
\label{tab:pipllm-ai2thor}
\resizebox{\textwidth}{!}{
\setlength{\tabcolsep}{6pt}
\begin{tabular}{l *{15}{c}}
\toprule
\multirow{2}{*}{Methods} &
\multicolumn{5}{c}{Compound} &
\multicolumn{5}{c}{Complex} &
\multicolumn{5}{c}{Vague} \\
\cmidrule(lr){2-6} \cmidrule(lr){7-11} \cmidrule(lr){12-16}
& SR $\uparrow$ & \small{TC max/avg} $\downarrow$ & GCR $\uparrow$ & RU $\uparrow$ & Exe $\uparrow$ & SR $\uparrow$  & \small{TC max/avg} $\downarrow$ & GCR $\uparrow$ & RU $\uparrow$ & Exe $\uparrow$ & SR $\uparrow$& \small{TC max/avg} $\downarrow$ & GCR $\uparrow$ & RU $\uparrow$& Exe $\uparrow$\\
\midrule
CoT (GPT-4o) & 0.37 & 6.5/5.8 & 0.42 & 0.72 & 0.67 
& 0.00 & --/-- & 0.16 & 0.00 & 0.45 
& 0.00 & --/-- & 0.00 & 0.00 & 0.00 \\
SMART-LLM (GPT-4o) & 0.71 & 6.3/5.3 & 0.82 & 0.78 & 0.93 
& 0.26 & 7.3/6.0 & 0.39 & 0.63 & 0.75 
& 0.00 & --/-- & 0.12 & 0.00 & 0.67 \\
LaMMA-P (GPT-4o) & \textbf{0.93} & 6.3/5.0 & 0.94 & 0.91 & 0.92 
& 0.78 & 7.0/5.8 & 0.88 & 0.87 & \textbf{1.00} 
& 0.52 & 7.0/5.5 & 0.56 & 0.71 & \textbf{0.91} \\
PIP-LLM (ours, GPT-4o)   & \textbf{0.93} & \textbf{5.3/4.3} & \textbf{0.95} & \textbf{1.00} & \textbf{1.00} 
& \textbf{0.87} & \textbf{5.8/4.5} & \textbf{0.92} & \textbf{0.90} & \textbf{1.00} 
& \textbf{0.68} & \textbf{6.0/4.8} & \textbf{0.77} & \textbf{0.82} & \textbf{0.91} \\
PIP-LLM (ours, Qwen-32B)  & 0.63 & 5.8/4.8 & {0.72} & \textbf{1.00} & {0.93} 
& 0.48 & 6.5/5.5 & 0.45 & 0.81 & 0.64 
& 0.40 & 0.14/0.14 & 0.35 & 0.77 & 0.54 \\
\bottomrule
\end{tabular}
}
\end{table*}

\begin{table}[t]
\centering
\small
\resizebox{0.5\textwidth}{!}{
\begin{tabular}{l cc cc cc}
\toprule
\multirow{2}{*}{Methods} & \multicolumn{2}{c}{Compound} & \multicolumn{2}{c}{Complex} & \multicolumn{2}{c}{Vague} \\
\cmidrule(lr){2-3}\cmidrule(lr){4-5}\cmidrule(lr){6-7}
& SR & TC max/avg & SR & TC max/avg & SR & TC max/avg \\
\midrule
PIP-LLM w/o $\mathcal{D}$              & 0.49          & {5.8/5.0}                & 0.35          & 5.8/4.8               & 0.16          & 6.8/5.3\\
PIP-LLM w/o $\mathcal{B}\&\mathcal{V}$ & 0.68          & 5.5/4.3                  & 0.43      & 6.5/5.0               & 0.32          & 6.3/4.8 \\
PIP-LLM w/ $\mathcal{N}=1$             & 0.85          & {5.5/4.8}                & 0.74        & 5.3/4.5                & 0.44          & 6.0/4.8\\
PIP-LLM w/ $\mathcal{N}=3$             & {0.90}        & 5.8/4.8                  & 0.83         & 6.0/4.8                & 0.60          & 6.0/4.5 \\
PIP-LLM w/ $\mathcal{N}=4$             & \textbf{0.93} & \textbf{5.3/4.3}    & \textbf{0.87} & \textbf{5.8/4.5} & \textbf{0.68} & \textbf{6.0/4.8} \\
PIP-LLM w/o $\mathcal{AS}$             & 0.93          & 6.5/5.3                  & 0.87          & 7.0/5.8               & 0.68          & 7.3/6.5 \\
\bottomrule
\end{tabular}
}
\caption{Ablations across three categories. We consider the following key components in Fig. \ref{fig:framework}: team-level planning PDDL domain ($\mathcal{D}$), debugger ($\mathcal{B}$), verifier ($\mathcal{V}$),  the number of allowed refining loops ($\mathcal{N}$), and the task allocation and scheduling component ($\mathcal{AS}$). Evaluation is based on GPT-4o.}
\label{tab:ablation_sr_tc_only}
\vspace{-10pt}
\end{table}

\subsection{AI2THOR Benchmark Dataset}
To evaluate the performance of PIP-LLM and facilitate comparisons with baseline methods, we created a benchmark dataset by inheriting and modifying those from~\cite{zhang2024lamma, kannan2024smart}. Specifically, the dataset consists of three categories of tasks to better evaluate the performance of the proposed framework with an increasing level of complexity. 
\begin{itemize}
    \item \textbf{Compound Tasks} involve multiple objects and can be decomposed into sequential or parallel subtasks. The robot team is homogeneous, and each robot has the necessary skills to finish all tasks. We need to properly allocate and schedule tasks among robots. For example, \texttt{Slice the lettuce, trash the mug and switch off the light}. 
    \item \textbf{Complex Tasks} are intended for heterogeneous robot teams. It resembles Compound Tasks in their characteristics, like task decomposition, multi-robot engagement, and the presence of multiple objects. However, unlike Compound Tasks, where individual robots can perform subtasks independently, robots may need to perform a subtask cooperatively due to limitations in their skills, e.g., a robot does not have the skill to open the fridge/microwave; a robot can only afford one unit of weight and cannot move an object of two units of weight\footnote{AI2THOR does not support cooperative manipulation. When dealing with such tasks, we manually inspect the plan and cost output from frameworks, but do not visualize plans in the simulator. }. We create a set of such tasks by either setting the robots' corresponding skill to zero or increasing the weight of objects. Such tasks require strategic team coordination for effective task completion. 
    \item \textbf{Vague Tasks} These tasks present additional challenges with ambiguous natural language instructions, which require the robots to infer missing details. For example, \texttt{gather up 3 school supplies on the bed}; \texttt{It is very bright and we do not need light}. 
\end{itemize}
Our dataset includes 41 compound tasks, 23 complex tasks,  and 25 vague command tasks to evaluate task decomposition, allocation, and execution efficiency.

\subsection{Evaluation Baselines and Metrics}
We use the following evaluation metrics: Success Rate (SR), Robot Utilization (RU), Goal Condition Recall (GCR), Executability (Exe), Sum of Travel Cost (TC-sum), Max of Travel Cost (TC-max), following~\cite{kannan2024smart, zhang2024lamma, singh2023progprompt}. Our evaluations are based on the dataset’s final ground truth demonstrated by humans. 
\begin{itemize}
     \item \textit{Exe} is the fraction of actions in the task plan that can be
     executed, regardless of their impact on task completion.
      \item \textit{RU} measures the planning efficiency of the action sequence by comparing the total transitions from all successful executions to the overall corresponding ground truth count.
     \item \textit{GCR} is computed using the set difference between
     ground truth goal conditions and final conditions
     achieved, divided by the total number of ground truth goal conditions.
     \item \textit{SR} is calculated as the ratio of successful executions to the total number of tasks.
       \item \textit{TC-avg} is the average travel cost for robots to finish all the tasks.
       It is a metric for the coordination efficiency of the team.
       \item \textit{TC-max} is the maximum total travel cost for robots to finish all the tasks.
       It is a metric for the team's load-balancing efficiency.
\end{itemize}

We use SMART-LLM~\cite{kannan2024smart}, LaMMA-P~\cite{zhang2024lamma}, and a basic Chain of Thought (CoT) as our baselines using GPT-4o. We evaluate our framework PIP-LLM using both  GPT-4o and Qwen-32B.

\begin{figure*}[ht]
    \centering
    \subfloat[ours:PIP-LLM]{
    \includegraphics[width=0.23 \textwidth]{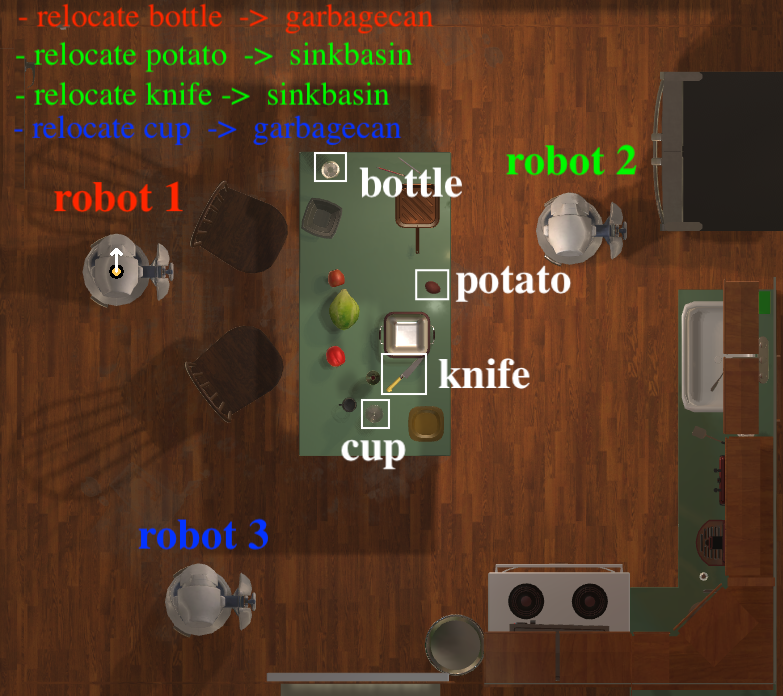}
    \label{fig:step21}
    } 
    \subfloat[]{
    \centering
    \includegraphics[width=0.23 \textwidth]{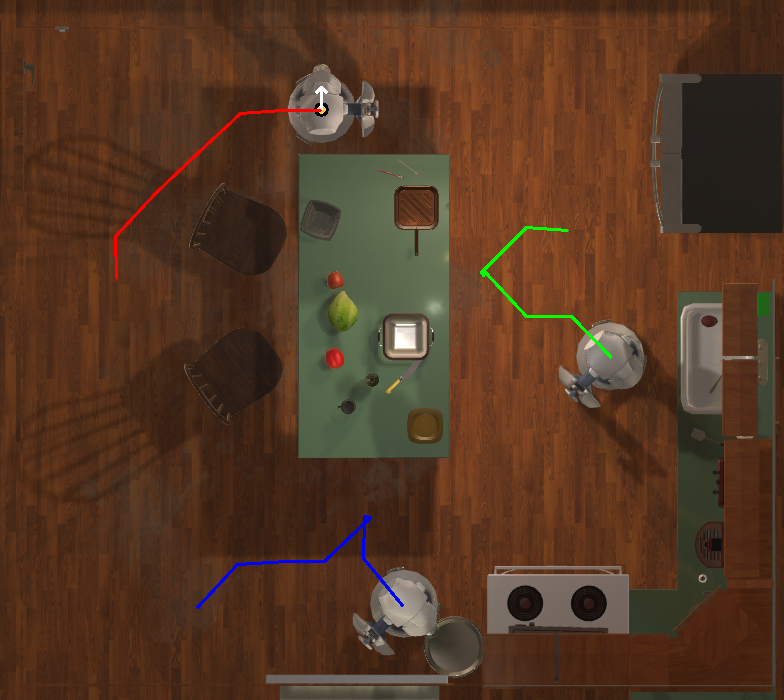}
    \label{fig:step31}
    }
    \subfloat[]{
    \centering
    \includegraphics[width=0.23 \textwidth]{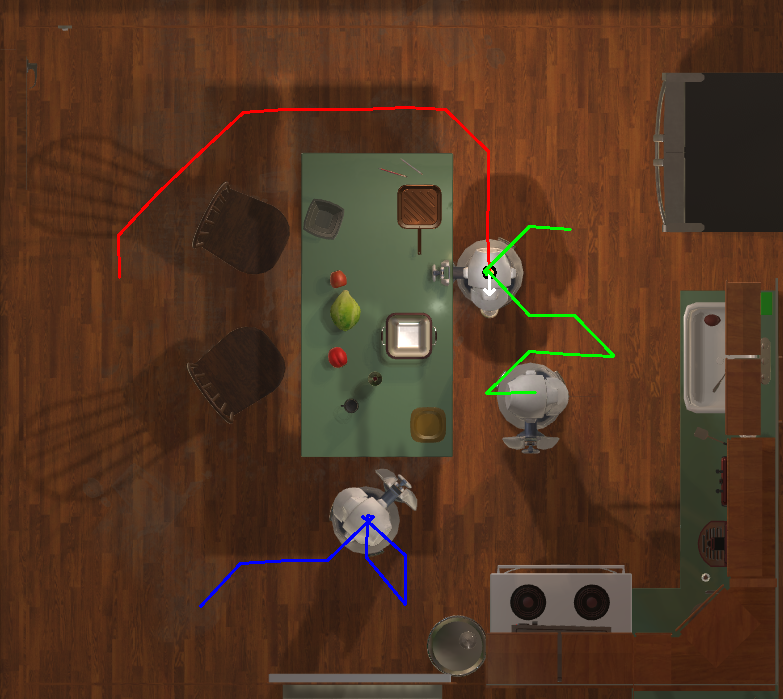}
    \label{fig:step41}
    }
    \subfloat[]{
    \centering
    \includegraphics[width=0.23 \textwidth]{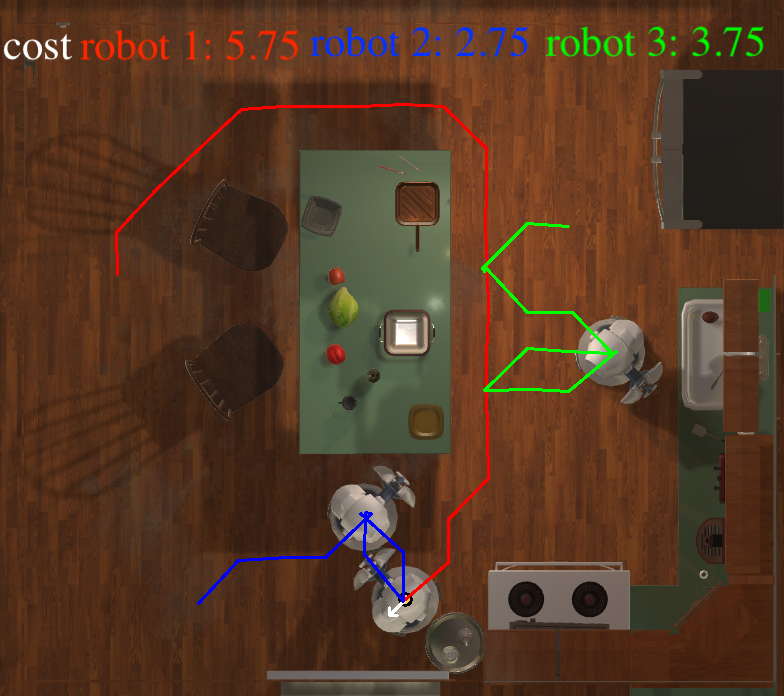}
    \label{fig:step51}
    } \\
    \subfloat[SMART-LLM \& LaMMA-P]{
    \includegraphics[width=0.23 \textwidth]{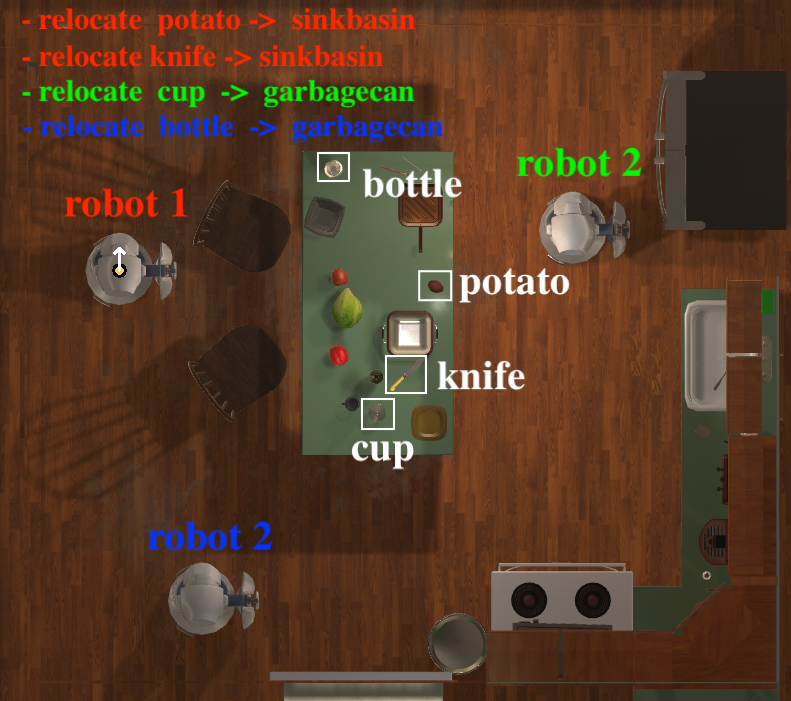}
    \label{fig:base_step11}
    }
    \subfloat[]{
    \includegraphics[width=0.23 \textwidth]{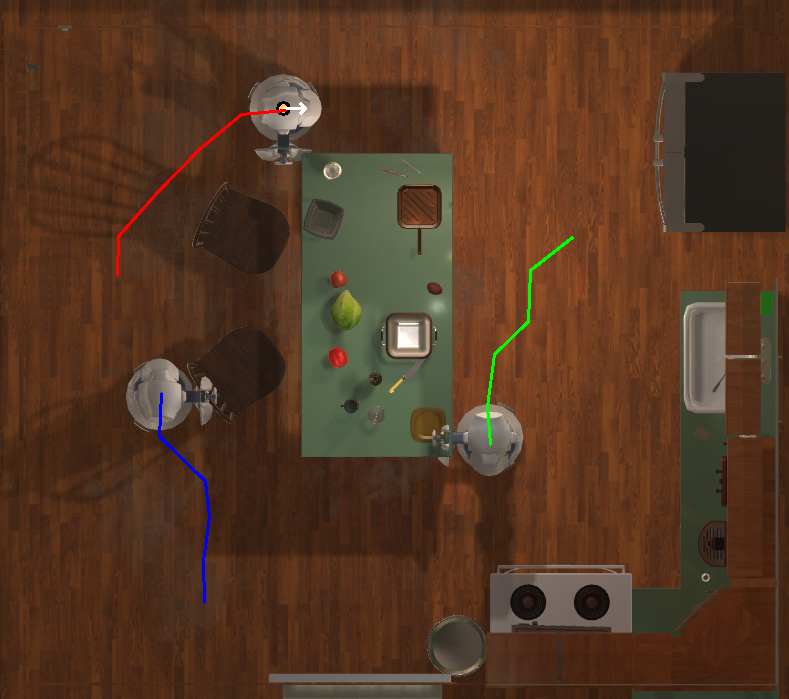}
    \label{fig:base_step21}
    } 
    \subfloat[]{
    \centering
    \includegraphics[width=0.23 \textwidth]{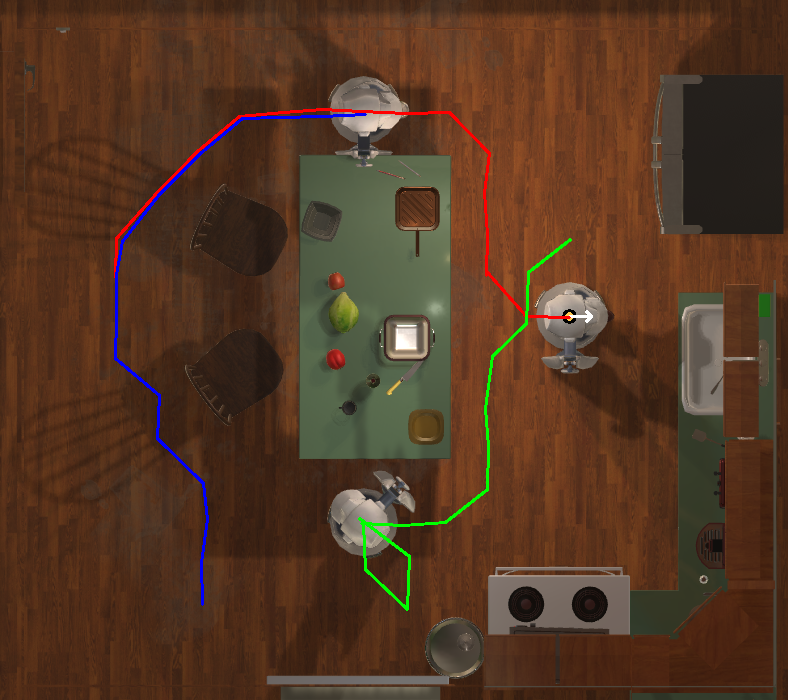}
    \label{fig:base_step31}
    }
     \subfloat[]{
    \centering
    \includegraphics[width=0.23 \textwidth]{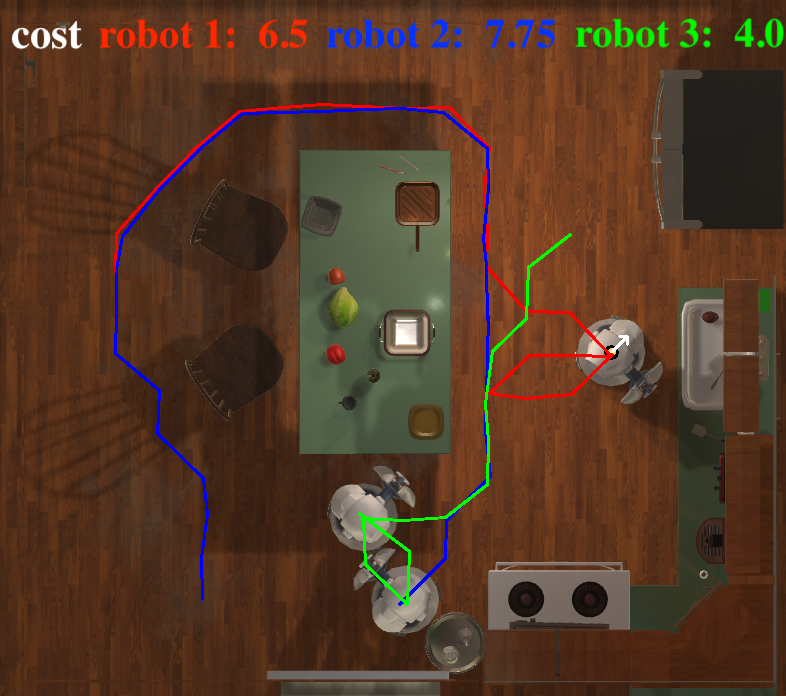}
    \label{fig:base_step51}
    }
    \caption{
     An illustrative example to show the differences between our framework, PIP-LLM, and baselines (SMART-LLM and LaMMA-P). The task is \texttt{put potato and knife in sinkbasin and trash the bottle and cup.} (a)-(d) show the execution of results from PIP-LLM. (e)-(h) show the execution of baselines. The top left of (a) and (e) shows the allocated tasks of all robots. The top of (d) and (h) shows the cost of all robots. Ours achieves much smaller maximum/average cost. Red, blue, and green colors correspond to robots 1, 2, and 3, respectively. }
    \label{fig:qualitative_example}
    \vspace{-15pt}
\end{figure*} 

\subsection{Results for AI2-THOR Environment}
\subsubsection{Qualitative results} An illustrative example is shown in Fig. \ref{fig:qualitative_example}. The task description is given in the caption. The first row presents screenshots of ours during the task execution, and the second row corresponds to baselines. The travel cost of each robot is shown in Fig. \ref{fig:step51} and \ref{fig:base_step51}, respectively. As we can see from the figures, our approach, by using object information and robot cost to optimize the task allocation and scheduling, results in a smaller travel cost (shorter paths for each robot), i.e., better coordination efficiency.   

\subsubsection{Quantitative results} The main statistical results are shown in Tab. \ref{tab:pipllm-ai2thor}. Overall, in all GPT-4o-based frameworks, our framework shows significant improvement in travel cost-related metrics (TC-sum and TC-max) compared to the state-of-the-art baselines and shows a substantial increase in the success rate, especially in the vague tasks group. Such results suggest that our framework can not only improve the team coordination efficiency but also augment LLMs' capability to understand and analyze the tasks. We attribute such improvements to the following reasons. First, our framework introduces a hierarchical structure: team first, then robots. On the team level, LLMs will be guided by the PDDL domain and do not need to go into details of each robot, thus being less likely to be distracted by redundant information to hallucinate. Besides, we introduce the debugger and checker structure to allow the LLMs to correct themselves if the first round of generation is not right.  Second, on the robot level, we incorporate an explicit cost-aware task allocation and scheduling structure, which will inherently improve the coordination efficiency. Tab. \ref{tab:pipllm-ai2thor} also shows the result of using Qwen-32B for our framework. 
In such a case, the performance of our framework, in terms of both success rate and travel cost, remains significantly better than the naive CoT framework with GPT-4o. 
These results demonstrate the effectiveness of the proposed framework.


\subsection{Ablation Study for AI2-THOR Environment}
We conducted an ablation study to assess the contribution of individual components to the performance of PIP-LLMs in Tab. \ref{tab:ablation_sr_tc_only}. The following key components were considered: the team-level planning PDDL domain ($\mathcal{D}$), the debugger ($\mathcal{B}$), the verifier ($\mathcal{V}$), the number of allowed refinement loops ($\mathcal{N}$), and the task allocation and scheduling module ($\mathcal{AS}$).
In the first row, we exclude the PDDL domain $\mathcal{D}$ and allow the LLMs to design team-level actions autonomously, performing planning without generating explicit PDDL problems. As shown in the table, the success rate drops substantially across all three task categories, underscoring the critical role of the PDDL domain in guiding task understanding and decomposition.
In the second row, we remove the solution refinement loop, i.e., the debugger and verifier components ($\mathcal{B}$ and $\mathcal{V}$). This results in a pronounced decline in success rate, particularly for the complex and vague task groups, highlighting the effectiveness of $\mathcal{B}$ and $\mathcal{V}$. To further examine their impact, we vary the number of refinement loops ($\mathcal{N}$) from the third to the fifth rows. The results clearly indicate that increasing $\mathcal{N}$ consistently improves performance. We adopt $\mathcal{N}=4$ for the results reported in Tab.~\ref{tab:pipllm-ai2thor}.
Finally, in the last row, we replace our task allocation and scheduling module with a purely LLM-based approach. 
This modification led to a substantial increase in travel-related costs, demonstrating the efficiency of the proposed $\mathcal{AS}$ component in optimizing resource utilization.

\subsection{Results for Gazebo Warehouse Environment}

\begin{figure}
    \centering
    \includegraphics[width=0.98\linewidth]{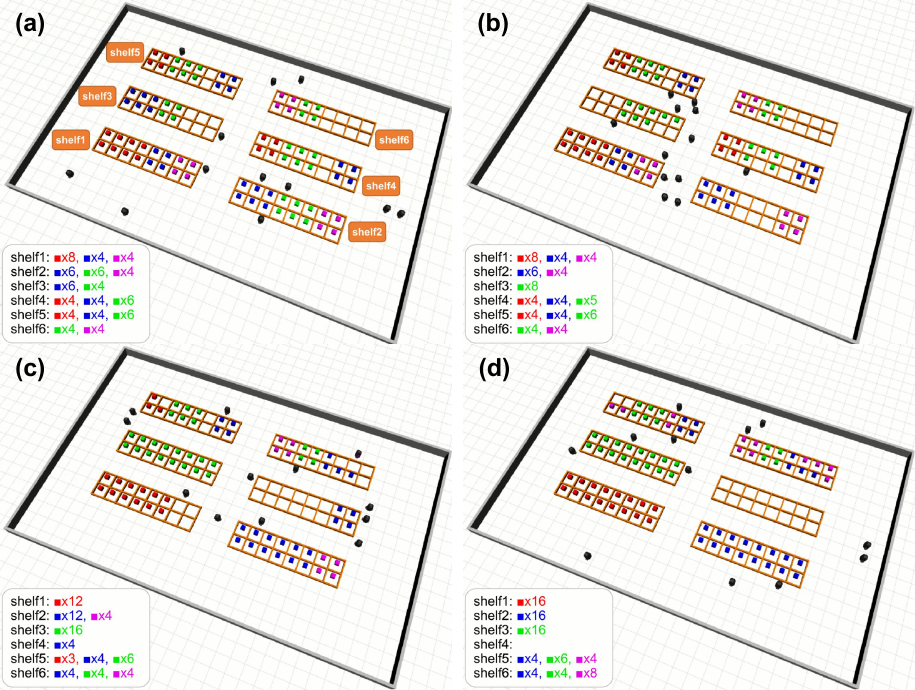}  
    \caption{A team of 12 robots conducting tasks in the warehouse. Products 1, 2, 3, and 4 are in red, blue, green, and magenta, respectively. The task command is \texttt{There should be full of product 1 in shelf 1; full of product 2 in shelf 2; full of product 3 in shelf 3; and shelf 4 should be empty. } The team-level action plan is \texttt{0.0: (move-product shelf2 shelf3 product3 magnitude6)  1.0: (move-product shelf3 shelf2 product2 magnitude6), ..., 8.0: (move-product shelf4 shelf2 product2 magnitude4)} (a)-(d) show the changes in product locations over time.}
    \label{fig:gazebo_exp}
    \vspace{-3pt}
\end{figure}

\begin{figure}
    \centering
    \includegraphics[width=0.93\linewidth]{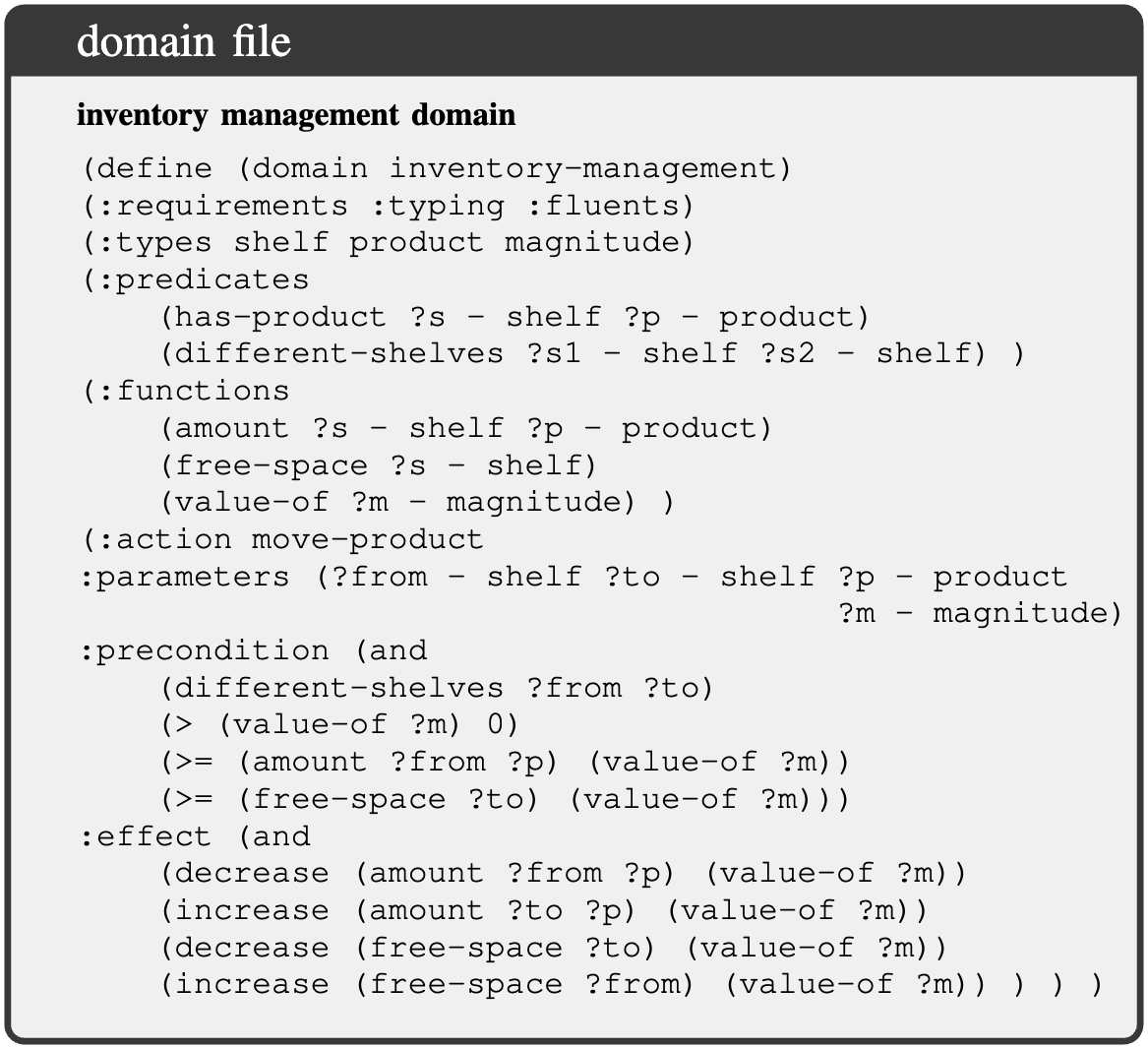}    
    \caption{PDDL domain used for the Gazebo warehouse environment.}
    \label{fig:pddl_domain}
    \vspace{-15pt}
\end{figure}

Since AI2THOR is designed for a small-scale household setup, we cannot use it to test scalable multi-robot coordination. To this end,  we create a customized warehouse environment in Gazebo for the second part of our experiments, as shown in Fig. \ref{fig:gazebo_exp},  in which a team of robots needs to move many objects between different shelves based on the input of language instructions. We use blocks of different colors to represent different objects that need to be moved. We assume that the states of robots and objects are known. The team-level planning PDDL domain file is shown in Fig. \ref{fig:pddl_domain}, which focuses on only moving products from one shelf to another shelf without referring to any specific robots. An illustrative task example is given in the caption of Fig. \ref{fig:gazebo_exp}. The PDDL plan, i.e., a sequence of team actions, will be generated and fed to the task allocation and scheduling part to decide which robots should be used to accomplish the team action. The team can efficiently rearrange a large number of objects.

Tab. \ref{tab:warehouse} shows ten representative tasks for the warehouse environment. 
A task is considered finished successfully if the product distribution on the shelves is the same as the ground truth annotated by a human. To estimate the success rate, we execute each task ten times from varied initial robot locations, explicitly accounting for GPT-4o’s stochastic output. 
As shown in Tab. \ref{tab:warehouse}, our framework is able to achieve a significantly higher success rate meanwhile inducing a much lower travel cost. Baselines include CoT, SMART-LLM, and LaMMA-P. We observe that these baselines succeed only in the first task. This is because in task 1, the initial product distribution is very close to the desired distribution: we just need to move two products 1 from shelf 1 to shelf 3. The baseline frameworks can manage such ``one-step" simple planning. For other tasks, they all involve moving products back and forth between shelves, which causes all baselines to fail.
These results indicate that our framework can effectively understand high-level instructions to coordinate a large number of robots with a high coordination efficiency.  Moreover, in tasks 6 and 7, we introduce noise to the language commands by using misspellings. We observe that such small noise can still lead the framework to fail occasionally. For example, in task 6,  product 3 is written as \textit{prodcut3} on purpose and LLMs may generate a PDDL problem by claiming \textit{prodcut3} as a new product instead of product 3. Similarly, LLMs may claim \textit{shelve4} as a new shelf in the generated PDDL file.

\begin{table}[t]
\centering
\caption{Statistical results from tabletop experiments.}
\label{tab:warehouse}
\setlength{\tabcolsep}{5pt}
\renewcommand{\arraystretch}{1.2}

\begin{tabular}{c c c c c}
\toprule
\multicolumn{1}{c}{\textbf{Task}} &
\multicolumn{2}{c}{\textbf{Success rate (\%)}} &
\multicolumn{2}{c}{\textbf{Travel cost} ($m$)} \\
\cmidrule(lr){1-1}\cmidrule(lr){2-3}\cmidrule(lr){4-5}
\textbf{ID} & \textbf{ours} & \textbf{baselines} & \textbf{ours} (max/avg) & \textbf{baselines}  \\
\midrule
1 & 100 & 100 & $28.5/33.0$ & $34.1/40.3$  \\
2 & 100 & 0 & $164.7/141.1$ &  \NA  \\
3 & 100 & 0 & $151.2/111.6$ & \NA  \\
4 & 100 & 0 & $321.3/267.1$ & \NA \\
5 & 90 & 0 & $194.3/167.1$ & \NA  \\
6 & 80 & 0 & $201.4/157.4$ & \NA  \\
7 & 70 & 0 & $304.2/271.4$ & \NA  \\
8 & 100 & 0 & $111.6/97.4$ & \NA \\
9 & 100 & 0 & $221.2/177.4$ & \NA  \\
10 & 100 & 0 & $288.3/248.3$ & \NA  \\
\bottomrule
\end{tabular}

\vspace{0.8em}

\begin{tabularx}{\linewidth}{c X}
\toprule
\textbf{ID} & \textbf{Task description} \\
\midrule
1 & There should be 7 product 1 in shelf 3. \\
2 & There should be 3 product2 in shelf4, 5 product2 in shelf2. \\
3 & There should be 10 product1 in shelf3, 9 product1 in shelf2. We can only use robot1 to move product to [location] of the shelf1 \\
4 & There should be 2 product2, 7 product3, 8 product1 in shelf3, 3 product2, 5 product3, 12 product1 in shelf2. \\
5 & There should be 20 product1 in shelf2, 16 product1 in shelf4, 2 product1 in shelf3. But robot 0 should not be used to move product1 from one shelf to another. \\
6. & On shelf4, there oughta be 14 prodcut3, 2 prodcut1, and 1 prduct2. (\textit{include misspellings by design})\\
7. & Theyre supposed to be 17 prodct1 on shelve4, I think. \\
8. & shelf4 has ten plus seven product1 \\
9. & shelf2 stocks a quartet of product2 and a trio of product3\\
10. & shelf4 stores a dozen product3 and a quintet of product1. \\
\bottomrule
\end{tabularx}
\vspace{-7pt}
\end{table}

\begin{figure}
\vspace{0.2cm}
\centering
\includegraphics[width=0.48\textwidth]{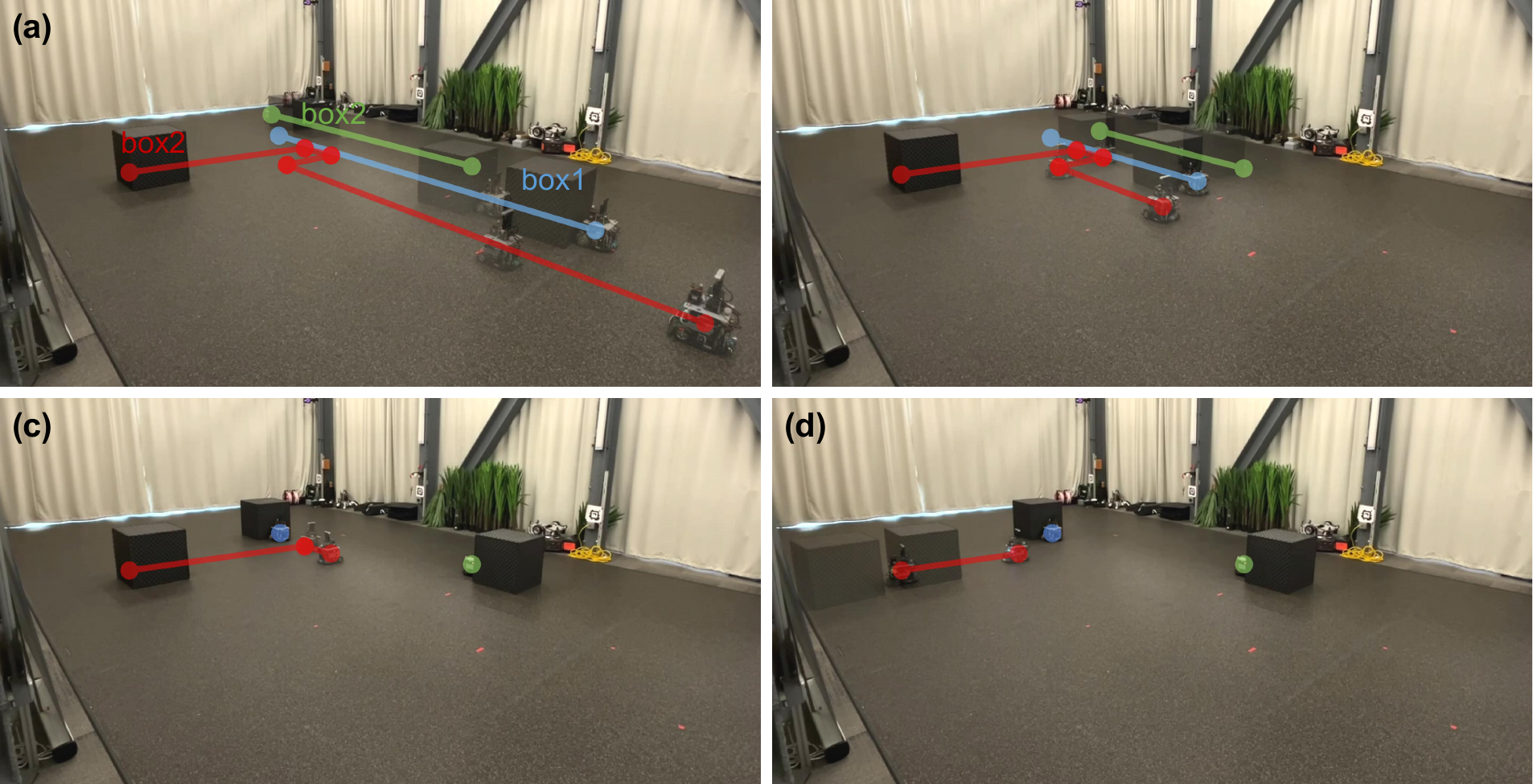}
\caption{Snapshots of hardware experiments. The task is \texttt{Push boxes to their corresponding goals.} The planned paths for robots are marked as blue, green, and red dots, respectively, in (a). Three robots are available for the task. The transparent robots and boxes along the paths denote their future locations. }
\label{fig:hardware}
\vspace{-0.8cm}
\end{figure}

\noindent \textbf{Hardware Demonstration.} 
We employ multiple mobile ground robots to perform object transfer operations, where each robot is tasked with pushing boxes to designated goal locations.
Snapshots of a representative experiment at different time frames are shown in Fig. \ref{fig:hardware}.

\noindent \textbf{Limitations.} The proposed framework relies on a pre-defined team-level PDDL planning domain. Such a domain encapsulates all possible transitions in the environment and can effectively guide LLMs to find team-level solutions. Currently, it is still very challenging for LLMs to automatically design such a domain given a general problem context description. Another limitation is that we assume the world is closed and static, as in the classic AI planning. If there exists uncertainty in the environment (e.g., the number and type of objects are changing or the execution of actions succeeds with a probability), some replanning mechanism should be included to close the loop for PIP-LLM.

\section{Conclusion}\label{sec:conclusion}
This work presents a hierarchical framework, PIP-LLM, that integrates LLMs with both PDDL and IP to support scalable, long-horizon multi-robot task planning. By leveraging the complementary strengths of PDDL and IP, and using LLMs to bridge natural language inputs with formal representations, the framework lowers the barrier to entry for non-experts while maintaining the rigor needed for complex problem-solving. Through comprehensive simulations and hardware validation, we demonstrate the framework's effectiveness and potential in enabling intuitive deployment of multi-robot coordination systems across diverse scenarios. The code and dataset will be released after acceptance.

\bibliographystyle{IEEEtran}
\bibliography{IEEEabrv, RAL2025}

\end{document}